\documentclass[conference,compsoc,a4paper]{IEEEtran}[2015/08/26]

\usepackage{pbalance}

\usepackage{upquote}

\usepackage[ngerman,main=english]{babel}
\addto\extrasenglish{\languageshorthands{ngerman}\useshorthands{"}}

\usepackage[hyphens]{url}

\makeatletter
\g@addto@macro{\UrlBreaks}{\UrlOrds}
\makeatother

\usepackage[zerostyle=b,scaled=.75]{newtxtt}

\usepackage[T1]{fontenc}

\usepackage[
  babel=true, 
  expansion=alltext,
  protrusion=alltext-nott, 
  final 
]{microtype}

\DisableLigatures{encoding = T1, family = tt* }

\usepackage{graphicx}

\usepackage{diagbox}

\usepackage{xcolor}

\usepackage{listings}

\definecolor{eclipseStrings}{RGB}{42,0.0,255}
\definecolor{eclipseKeywords}{RGB}{127,0,85}
\colorlet{numb}{magenta!60!black}
\newcommand{\skipping}[1]{}

\lstdefinelanguage{json}{
    basicstyle=\normalfont\ttfamily,
    commentstyle=\color{eclipseStrings}, 
    stringstyle=\color{eclipseKeywords}, 
    numbers=left,
    numberstyle=\scriptsize,
    stepnumber=1,
    numbersep=8pt,
    showstringspaces=false,
    breaklines=true,
    frame=lines,
    string=[s]{"}{"},
    comment=[l]{:\ "},
    morecomment=[l]{:"},
    literate=
        *{0}{{{\color{numb}0}}}{1}
         {1}{{{\color{numb}1}}}{1}
         {2}{{{\color{numb}2}}}{1}
         {3}{{{\color{numb}3}}}{1}
         {4}{{{\color{numb}4}}}{1}
         {5}{{{\color{numb}5}}}{1}
         {6}{{{\color{numb}6}}}{1}
         {7}{{{\color{numb}7}}}{1}
         {8}{{{\color{numb}8}}}{1}
         {9}{{{\color{numb}9}}}{1}
}

\usepackage[autostyle=true]{csquotes}

\defineshorthand{"`}{\openautoquote}
\defineshorthand{"'}{\closeautoquote}

\usepackage{booktabs}

\usepackage{paralist}

\usepackage[%
  square,        
  comma,         
  numbers,       
  sort&compress 
]{natbib}

\usepackage{etoolbox}
\makeatletter
\patchcmd{\NAT@test}{\else \NAT@nm}{\else \NAT@hyper@{\NAT@nm}}{}{}
\makeatother

\usepackage{pdfcomment}

\usepackage{stfloats}
\fnbelowfloat

\usepackage[group-minimum-digits=4,per-mode=fraction]{siunitx}
\addto\extrasgerman{\sisetup{locale = DE}}

\usepackage{footmisc}

\usepackage{hyperxmp}

\usepackage{hyperref}

\hypersetup{
  keeppdfinfo,
  hidelinks,
  colorlinks=true,
  allcolors=black,
  pdfstartview=Fit,
  breaklinks=true,
  pdftitle={High-Resolution Multi-Scale RAFT},
  pdfsubject={Technical Report for Robust Vision Challenge 2022},
  pdfauthor={Azin Jahedi, Maximilian Luz, Lukas Mehl, Marc Rivinius, Andrés Bruhn},
  pdfkeywords={
    Robust Vision Challenge 2022,
    Optical Flow,
    Multi-scale,
    Coarse-to-fine%
  },
}

\usepackage[all]{hypcap}

\usepackage[caption=false,font=footnotesize]{subfig}

\usepackage[capitalise,nameinlink,noabbrev]{cleveref}

\crefname{listing}{Listing}{Listings}
\Crefname{listing}{Listing}{Listings}
\crefname{lstlisting}{Listing}{Listings}
\Crefname{lstlisting}{Listing}{Listings}

\usepackage{lipsum}

\usepackage[math]{blindtext}
\usepackage{mwe}
\usepackage[realmainfile]{currfile}
\usepackage{tcolorbox}
\tcbuselibrary{listings}

\DeclareFontFamily{U}{MnSymbolC}{}
\DeclareSymbolFont{MnSyC}{U}{MnSymbolC}{m}{n}
\DeclareFontShape{U}{MnSymbolC}{m}{n}{
  <-6>    MnSymbolC5
  <6-7>   MnSymbolC6
  <7-8>   MnSymbolC7
  <8-9>   MnSymbolC8
  <9-10>  MnSymbolC9
  <10-12> MnSymbolC10
  <12->   MnSymbolC12%
}{}
\DeclareMathSymbol{\powerset}{\mathord}{MnSyC}{180}

\usepackage{xspace}

\makeatletter
\newcommand{\hydash}{\penalty\@M-\hskip\z@skip}
\makeatother

\input glyphtounicode
\IEEEoverridecommandlockouts

\begin{document}

\title{High-Resolution Multi-Scale RAFT \\ (Robust Vision Challenge 2022)}

\author{%
  \IEEEauthorblockN{Azin Jahedi\IEEEauthorrefmark{1}, Maximilian Luz\IEEEauthorrefmark{1}, Lukas Mehl\IEEEauthorrefmark{1}, Marc Rivinius\IEEEauthorrefmark{2}, Andrés Bruhn\IEEEauthorrefmark{1}}
  \IEEEauthorblockA{\IEEEauthorrefmark{1}Institute for Visualization and Interactive Systems, University of Stuttgart, Germany}
  \IEEEauthorblockA{\IEEEauthorrefmark{2}Institute of Information Security, University of Stuttgart, Germany}

   \thanks{The authors thank the Deutsche Forschungsgemeinschaft (DFG, German Research Foundation) -- Project-ID 251654672 -- TRR 161 (B04) for funding and the International Max Planck Research School for Intelligent Systems (IMPRS-IS) for supporting Azin Jahedi.}
}


\maketitle

\thispagestyle{plain}
\pagestyle{plain}

%
%

\begin{abstract}%
\skipping{In this report we present details of our optical flow approach MS-RAFT+ that won the Robust Vision Challenge 2022. Our approach is based on the MS-RAFT method which successfully integrates several multi-scale concepts into single-scale RAFT. 
To improve accuracy, we extend MS-RAFT's multi-scale architecture by an additional finer scale at half the original resolution. 
Moreover to improve the generalization performance, we adjust the fine-tuning process.
The evaluation of the resulting MS-RAFT+ method makes the benefits of these adaptations explicit. Among all participating methods it ranks first on VIPER and second on KITTI 2015, Sintel and Middlebury resulting in the first place in the overall ranking of the Robust Vision Challenge.
}%
In this report, we present our optical flow approach, MS-RAFT+, that won
the Robust Vision Challenge 2022. It is based on the MS-RAFT method, which successfully integrates several multi-scale concepts into single-scale RAFT. 
Our approach extends this method
by exploiting an additional finer scale for estimating the flow, which is made feasible by on-demand cost computation. This way, 
it
can not only operate at half the original resolution, but also use MS-RAFT's shared convex upsampler to obtain full resolution flow. 
Moreover, our approach relies on an adjusted fine-tuning scheme during training.
This in turn 
aims at improving the generalization across benchmarks.
Among all participating methods in the Robust Vision Challenge, 
our approach ranks first on VIPER and second on KITTI, Sintel, and Middlebury, resulting
in the first place of the overall ranking.
\end{abstract}

\IEEEpeerreviewmaketitle

\section{Introduction}
\label{sec:introduction}
The estimation of the pixel-wise motion between two consecutive frames of an image sequence, the so-called optical flow, is a fundamental task in computer vision. Solving this task is useful for numerous applications ranging from action recognition and motion capture over tracking and surveillance to video processing and autonomous driving. This diversity of applications makes explicit that one desired property of optical flow algorithms is the generalizability across data sets. 
To evaluate this property, the Robust Vision Challenge was designed, requiring the development of algorithms that perform well on various benchmarks using the same model, parameters, and setup.
\section{Approach}
\label{sec:Approach}
Our method for the Robust Vision Challenge 2022 is based on our recent MS-RAFT approach, which combines four hierarchical concepts in one  estimation framework: (i) a coarse-to-fine computation scheme that provides useful initialization from coarser scales, (ii) a U-Net-style feature extractor which relies on semantically enhanced multi-scale features, (iii) 
RAFT's original correlation pyramid that introduces non-local cost information in the matching process, and (iv) a multi-scale multi-iteration loss that considers
a sample-wise robust loss function for fine-tuning the network.
We refer the reader to \cite{Jahedi2022_MSRAFT} for more details. 
For the Robust Vision Challenge, we extended the architecture of MS-RAFT by an additional finer scale, which is realized by an on-demand computation of the matching costs, and an adjusted fine-tuning step combining MS-RAFT's sample-wise robust loss function with a different mixed training strategy for better generalization. In the following, we discuss the details of these improvements.
\subsection{Architecture}
\label{subsec:architecture}
\begin{figure}
    \centering
       \includegraphics[width=1\columnwidth]{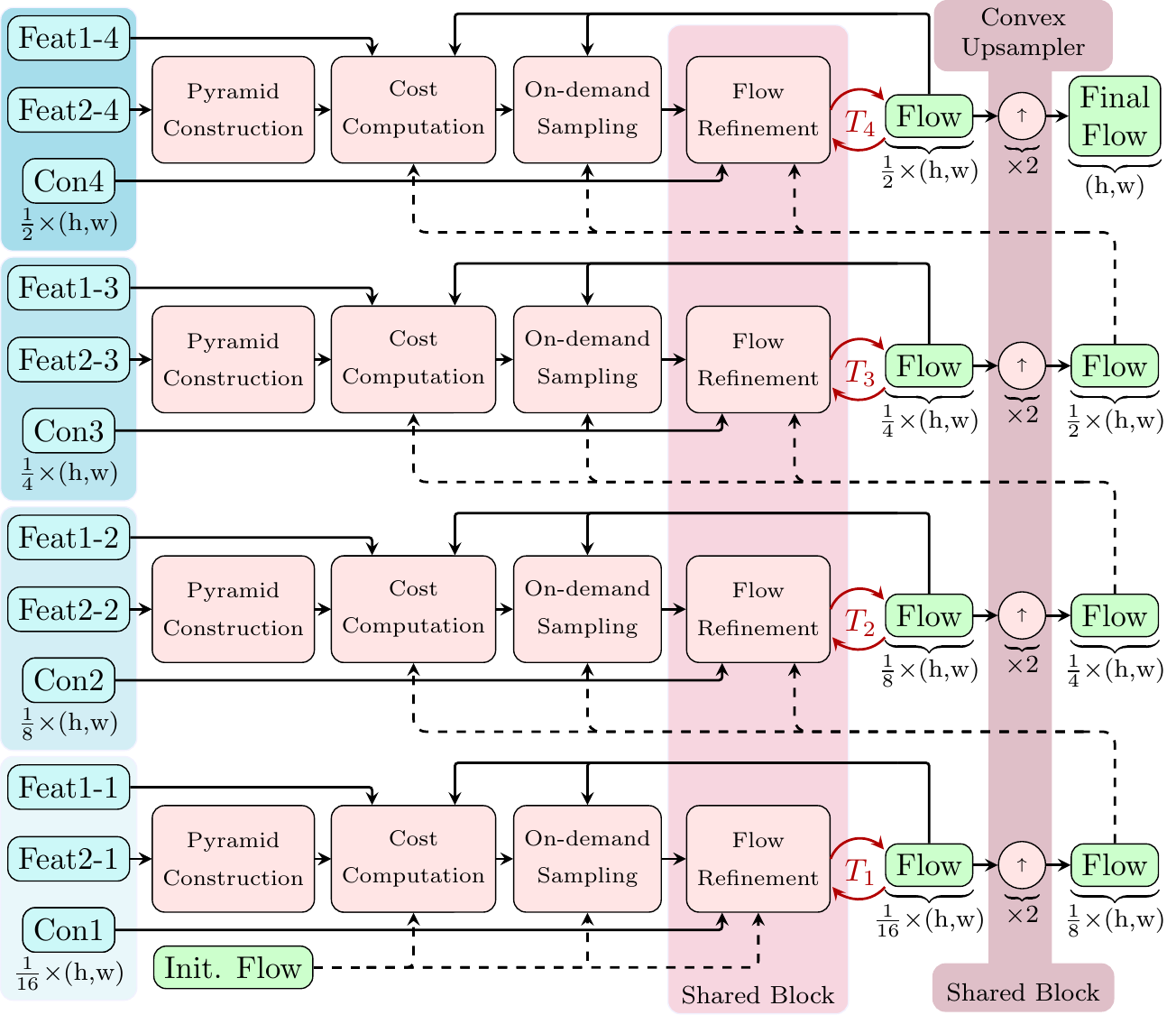}
    \caption{Our coarse-to-fine scheme. Best viewed as PDF} 
    \label{fig:ctf-flow-estimation}
\end{figure}
Our coarse-to-fine estimation scheme with four scales is shown in 
\cref{fig:ctf-flow-estimation}. Feat1-X, Feat2-X, ConX (shown in blue windows) are the inputs of this estimation sub-network, which represent image features (of both frames) and context features (of only the first frame) at different scales.
The spatial size of image and context features increases from bottom to top.
These feature volumes are computed via our U-Net-style feature encoder introduced in MS-RAFT, which we extended to include an additional finer scale. As shown in 
\cite{Jahedi2022_MSRAFT}
considering proper higher resolution inputs for estimating the flow together with a coarse-to-fine strategy improves the accuracy and allows the preservation of more structural details.
Therefore, we also extended MS-RAFT's coarse-to-fine scheme to contain an additional finer scale to gain further improvements. 

Given the aforementioned image and context features at different scales (resolutions), the flow is first estimated at the coarsest scale $S_1$. Based on an initial flow, which is typically set to zero if unavailable, the correlation costs between features of the first and the second frame are computed for the considered correspondence candidates;
cf.\ RAFT \cite{Teed2020_RAFT}.
These costs together with the current flow estimate and the context features are passed through a recurrent unit, which computes a flow update. After $T_1$ recurrent iterations at the coarsest scale, we perform a $\times 2$ convex upsampling. The resulting flow is then used as initialization at the next finer scale $S_2$. This procedure is repeated until the half-resolution flow has been computed at the fourth scale $S_4$. This flow is finally interpolated to the full resolution using $\times 2$ convex upsampling.
Importantly, the recurrent flow refinement sub-network and the $\times 2$ learned convex upsampler are shared among scales. Moreover, in contrast to MS-RAFT, no bilinear interpolation is needed in the full-resolution upsampling step.

Note that unlike RAFT and MS-RAFT, MS-RAFT+ does not pre-compute the all-pairs cost volume due to the extensive memory overhead imposed by the finer scales. Instead, the correlation costs of match candidates are computed on-demand during training and inference\footnote{RAFT \cite{Teed2020_RAFT} released the code for on-demand cost computation for inference. For training the corresponding modules needed completion and adjustments.}. 
Without the pre-computed all-pairs cost volume, one cannot pre-compute the correlation pyramid.
Instead, a pyramid of the second frame's features (Feat2-X in \cref{fig:ctf-flow-estimation}) is constructed via successive pooling operations.
A specialized CUDA kernel is then used to compute the desired costs directly:
It first computes correlation costs via a dot-product between first- and the pooled second-frame image features in a local neighborhood based on the current flow estimate.
Thereafter, it samples from this cost window to obtain the required cost values at the sub-pixel lookup positions.
Due to this modification, the extended coarse-to-fine scheme 
with the additional finer scale did not need extra memory for training or inference. However, both training and inference time were increased.

\subsection{Training}
\label{subsec:training}
Our MS-RAFT+ method with four scales has 16M parameters.
Thereby, the parameter overhead compared to the 5.3M parameters of RAFT and the 13.5M parameters of MS-RAFT is only due to the (extended) multi-scale U-Net-style feature extractor.
To train our method we used two Nvidia A100 GPUs with 40GB VRAM each.
Thereby, we applied $T_1=4$, $T_2=5$, $T_3=5$ and $T_4=6$ recurrent iterations, respectively, from the coarsest to the finest scale.
The overall training process took
around six days, compared to the three-day training of MS-RAFT with the same hardware. The extended training time was mainly due to the on-demand computation of the matching cost, which memory-wise allowed us to use the 
fourth scale, i.e.,\ the additional finer scale.

We trained our method for the Robust Vision Challenge in three phases: First, we pre-trained the network on Chairs \cite{Dosovitskiy2015_FlowNet} and then on Things \cite{Mayer2016_FTH}, each for 100K iterations using batch sizes of 10 and 6, respectively. 
Afterwards, we fine-tuned the network for another 100K iterations with batch size of 6 using a combination of datasets, where 30$\%$ of the training samples consist of Sintel \cite{Butler2012_Sintel}, 30$\%$ of Viper (every $10$th frame) \cite{Richter2017_VIPER}, 28$\%$ of KITTI 2015 \cite{Menze2015_KITTI}, 10$\%$ of Things \cite{Mayer2016_FTH} and 2$\%$ of HD1k \cite{Kondermann2016} with the same data augmentation as in \cite{Teed2020_RAFT}. 
Note that we did not use any samples from Middlebury \cite{Baker2011_MiddleburyFlow} for training, as the number of samples there is rather limited. 
The aforementioned proportions were found empirically by investigating different proportions for fine-tuning MS-RAFT, as its training is less time-consuming than MS-RAFT+'s training. The experiments for investigating the proportions were validated by evaluating the results on the Viper validation set (every 10th frame), a split of KITTI 2015, and Middlebury. As training supervision we used the multi-scale multi-iteration loss of MS-RAFT. As proposed in \cite{Jahedi2022_MSRAFT} we considered the L2 loss for pre-training and the robust sample-wise loss with $q=0.7$ for fine-tuning.
\subsection{Inference}
\label{subsec:inference}
Recent RNN-based methods typically use more iterations to refine the flow during inference than during training.
Furthermore, they often use a different number of iterations for each dataset to achieve better results.
For example, RAFT \cite{Teed2020_RAFT} was trained with 12 recurrent iterations, but applied 32 and 24 iteration for inference on Sintel and KITTI 2015, respectively. Similarly, also MS-RAFT used different numbers of iterations for Sintel and KITTI 2015.
In the case of MS-RAFT+, however, as we had to meet the requirements of the Robust Vision Challenge, we computed the optical flow using the same number of iterations for all benchmarks. To this end, we used $T_1=4$, $T_2=6$, $T_3=5$ and $T_4=10$ iterations from the coarsest to the finest scale, respectively, which were found to work best on the validation sets used previously in \cref{subsec:training} for determining the proportions of the mixed fine-tuning.

Regarding the flow initialization in coarse-to-fine schemes (cf. \cref{subsec:architecture}), typically a zero flow is used for initialization  at the coarsest level \cite{Sun2018_PWC}. However, as shown in single-scale RAFT \cite{Teed2020_RAFT}, the estimation can benefit from the predictions of the previous time-step by forward-warping the corresponding result and using it as improved initialization for the current time step. This so-called warm-start strategy used in \cite{Teed2020_RAFT} considers the forward-warped predicted flow from the previous time-step in a recursive manner, i.e.,\ the flow of the previous time step was also initialized by the forward-warped its previous pair's flow. 

During inference, we 
applied a cold warm-start strategy, in which only the previous image pair is considered for initialization. 
For example, to compute the flow between frame $I_3$ and frame $I_4$, we initialize the flow by forward-warping the flow between frame $I_2$ and frame $I_3$, for which, in turn, the zero flow was used as initialization.
In cases where the previous frame is not contained in the standard dataset files, or when the first image-pair in a sequence is considered, 
we use a zero-flow initialization---similar to the warm-start strategy in \cite{Teed2020_RAFT}.
\skipping{
\cref{tbl:coldwarm} shows the results of cold (always zero initialization) and cold warm initialization strategies for MS-RAFT+ on Sintel (Test) and Middlebury (Test) benchmarks. Note that among the target benchmarks, only these mentioned datasets provided the samples in sequences in the standard dataset files. Clearly, for the other datasets, as there was no previous samples available, zero initialization was used.
}

\section{Evaluation}
\label{subsec:evaluation}

\subsection{MS-RAFT+ vs. MS-RAFT}
Compared to the results of our MS-RAFT method  \cite{Jahedi2022_MSRAFT}, our new MS-RAFT+ approach leads to a significant improvement in accuracy for KITTI 2015 and Sintel (clean). Largest improvements can be observed in non-occluded regions (non-occluded Fl-all, EPE matched) as well as for small displacements (s0-10).
For Sintel (final), the overall results 
got slightly worse, but the accuracy improved for small displacements (s0-s10). The corresponding numbers can be found in \cref{eval:msraftplus_vs_msraft}.

\begin{table}[h!]
	\caption{Improvements of MS\_RAFT+ over MS\_RAFT}
	\label{eval:msraftplus_vs_msraft}
	\centering
	\begin{tabular}{l >{\hspace{-0.2cm}}c >{\hspace{-0.2cm}}c >{\hspace{-0.2cm}}c}
			\toprule
			 & {\bf \scriptsize MS\_RAFT}  &  {\bf \scriptsize MS\_RAFT+}  &  {\bf \scriptsize Improvement}  \\
			\midrule
			{\bf KITTI 2015 (Test)}  &  &  & \\[0.5mm]
			All Fl-all    &   4.88  &   {\bf 4.15}  &   {\bf +15.0\%} \\
			Non-occluded Fl-all  &  2.80  &    {\bf 2.18}& {\bf +22.1$\%$} \\
			\midrule
			{\bf Sintel Clean (Test)}   &  &    &   \\[0.5mm]
			EPE all      &     1.374 &  {\bf 1.232}   & {\bf +10.3\%} \\
		    EPE matched     & 0.479    &  {\bf 0.400}  &  {\bf +16.5\%} \\
		    	s0-10 & 0.221     &   {\bf 0.159}  & {\bf +28.1\%}
	 		\\
	 		
		    \midrule
			{\bf Sintel Final (Test)}   &    &    &   \\[0.5mm]
			
			EPE all    & {\bf 2.667}  & 2.682   &  \bf{-0.1}\% \\
			EPE matched  & \bf{1.190}  &  1.278   &   \bf{-6.8\%} \\
		    	s0-10  &  0.468   &   {\bf 0.420}   &  {\bf +10.2\%}
	 		\\
	 		\midrule
			{\bf Middlebury (Train)}   &     &     &  \\[0.5mm]
			EPE all      &  {0.184}$^*$   &   \bf{0.142}$^*$   &   \bf{+22.8}\% \\
		
			\bottomrule
		\end{tabular}
	
	\vspace{1mm}
	$^*$ Note that both methods were not trained on Middlebury.
\end{table}

\subsection{Cold warm-start vs. Cold-start}
When comparing the results of the cold warm-start initialization and the cold-start initialization for those benchmarks that provide entire sequences in the standard dataset files, we can observe larger improvements for Sintel (clean) and Sintel (final)---as in \cite{Teed2020_RAFT}---while the results for Middlebury hardly changed;
see \cref{tbl:coldwarm}. 
 
\begin{table}[h!]
	\caption{Cold-start vs. cold warm-start}
	\label{tbl:coldwarm}
	\centering
	\begin{tabular}{l >{\hspace{-0.2cm}}c >{\hspace{-0.2cm}}c >{\hspace{-0.2cm}}c >{\hspace{-0.2cm}}c}
			\toprule
			 & {\bf \scriptsize {Sintel Clean}}  &  {\bf \scriptsize Sintel Final}  &  {\bf \scriptsize Middlebury} & {\bf \scriptsize Middlebury} \\[-0.2mm]
			 & {\bf \scriptsize {(Test)}}  &  {\bf \scriptsize (Test) }  &  {\bf \scriptsize (Test)  } & {\bf \scriptsize (Test) } \\
			 & {\scriptsize EPE all}  &  {\scriptsize EPE all }  &  {\scriptsize EPE all} & {\scriptsize Avg-Rank } \\
			\midrule
		 {\bf {Cold}}   & 1.63 & 2.94 &  {\bf0.1725} & {\bf8.8}\\[0.5mm]
			{\bf Cold warm}    &  {\bf 1.23}  &   {\bf 2.68}  & {\bf 0.1725} & 9.2 \\

			\bottomrule
		\end{tabular}

\end{table}

\subsection{Robust Vision Challenge}
The results for the Robust Vision Challenge 2022 (RVC) are shown in \cref{fig:leaderboard}. Further details for the separate benchmarks, which we refer to in the following, can be found at the corresponding benchmark websites.

In the Middlebury benchmark, our MS-RAFT+ method achieves Rank 4 overall\footnote{\label{fn:rank}The overall rank also considers anonymous methods.} and Rank 2 among all RVC approaches. Since we did not use Middlebury training data for our mixed fine-tuning, this also shows the good generalization performance of our method.
In the RVC ranking for this benchmark, which also considers other metrics, 
it achieves a final rank of 2. 

In the KITTI 2015 benchmark, our method achieves Rank 8 (Fl-all) and Rank 2 (Fl-all non-occluded) overall\footref{fn:rank}, as well as Rank 2 (Fl-all) and Rank 1 (Fl-all non-occluded) among all RVC approaches. It is even on par with the leading scene flow method (CamLiFlow++) in non-occluded regions. 
In the RVC ranking for this benchmark, which also considers other metrics, our method achieves the 2nd place.

In the Sintel benchmark, MS-RAFT+ achieves overall\footref{fn:rank} Rank 19 (final, EPE all) and Rank 6 (clean, EPE-all), as well as Rank 3 (both final and clean, EPE-all) considering only RVC submissions.
Our method performs again particularly well in non-occluded regions (clean). Moreover, it shows very good results for small displacements (both final and clean), where it achieves 1st ranks in the corresponding metrics.
In the RVC ranking, 
that also considers many other metrics, our approach finally ranks 2nd.

Eventually, in the VIPER benchmark, our method achieves Rank 1 overall\footref{fn:rank} and Rank 1 among all RVC approaches. By clearly outperforming all other methods in all categories (day, sunset, rain, snow, night), it sets a new state-of-the-art for this benchmark. Consequently, in the final RVC ranking thereof, it ranks 1st.

With RVC-ranks of 2 for Middlebury, KITTI 2015, and Sintel, as well as 1 for VIPER, MS-RAFT+ shows a good generalization performance across all benchmarks. In the final leaderboard of the RVC, this results in Rank 1 of all participating methods (see \cref{fig:leaderboard}) and thereby our method won the Robust Vision Challenge 2022 for the optical flow estimation task.

\begin{figure}
    \centering
       \includegraphics[width=1\columnwidth]{rvc_small_cropped.pdf}
    \caption{Final leaderboard of Robust Vision Challenge 2022 (Top 5).} 
    \label{fig:leaderboard}
\end{figure}




\bibliographystyle{IEEEtranN} 
\bibliography{report}



\end{document}